\documentclass[letterpaper]{article} 
\usepackage{aaai2026}  
\usepackage{times}  
\usepackage{helvet}  
\usepackage{courier}  
\usepackage[hyphens]{url}  
\usepackage{graphicx} 
\graphicspath{{assets/rebuild_figures/}}
\usepackage{natbib}  
\usepackage{caption} 
\usepackage{amsmath}
\usepackage{amssymb}
\title{Evidence-Guided Unknown Rejection for High-Confidence Near-Known Unknowns}
\author{Xi Chen\textsuperscript{1}, Yingjun Xiao\textsuperscript{2}, Gang Fang\textsuperscript{3}}
\affiliations{
\textsuperscript{1}School of Computer Science and Cyber Engineering, Guangzhou University, Guangzhou 510006, China\\
\textsuperscript{2}School of Artificial Intelligence, Guangzhou University, Guangzhou 510006, China\\
\textsuperscript{3}Institute of Computing Science and Technology, Guangzhou University, Guangzhou 511370, China\\
Correspondence: gangf@gzhu.edu.cn
}

\begin{document}

\maketitle

\begin{abstract}
Open-set recognition systems face a neglected failure mode: high-confidence near-known unknowns, which lie outside the known label set but are close enough to known classes that a closed-set classifier accepts them with high confidence. We show that this failure is widespread across scalar-threshold methods, including recent post-hoc detectors, and that stronger encoders can amplify rather than remove the risk. We propose EGUR-A, which changes the decision from ``is this sample's score high enough?'' to ``does this predicted known class have sufficient evidence to accept this sample?'' EGUR-A combines class-conditional local acceptance evidence with global residual evidence, and selects their relative weight from known-sample statistics without unknown validation data. Across CUB, FGVC-Aircraft, and ImageNet-hard, EGUR-A substantially reduces high-confidence false known acceptance at matched known-rejection operating points. The result is not a stronger threshold; it is a different question: whether a known class is entitled to accept a sample.
\end{abstract}

\section{Introduction}

Open-set recognition (OSR) is usually framed as a thresholding problem: assign a label if a confidence or knownness score is high enough, and reject otherwise. This view hides a dangerous ambiguity. A bird outside the known taxonomy can look like a Nighthawk; an unseen aircraft variant can look like a C-47; at larger scale, a hard ImageNet unknown can share enough visual structure with a known object class to receive near-certain confidence. In such cases, the closed-set classifier may be highly confident, yet the predicted class may lack the evidence needed to accept the sample as a member. Classifier confidence and known-class acceptance evidence are not the same object.

The mechanism is structural. Softmax confidence is a relative score over the known label set: it measures which known class best explains a sample among the available labels, not whether any known class is entitled to accept it. Because the probabilities must sum to one, the classifier must allocate probability mass among known classes even when no known class is an adequate absolute match. When a near-known unknown falls inside the attraction basin of a known class, the closed-world classifier can concentrate probability mass on that class and saturate confidence toward 1. This only says that the sample wins the known-class competition for class $c$; it does not certify that the sample lies within the empirical support of class $c$.

This distinction matters most for \emph{near-known unknowns}: unknown samples that are semantically outside the known set but close to known classes in representation space. A high-confidence false acceptance is more harmful than an uncertain mistake because downstream systems may act on it without review, such as screening pipelines that escalate low-confidence cases but pass high-confidence ones through. Aggregate false known acceptance rate (FKAR) can hide this risk. On CUB, a DINOv2 linear MSP classifier has overall FKAR 0.163, yet its false known acceptance rate on unknown samples with closed-set confidence at least 0.9 is 0.442.

The problem is not solved by better features. Table \ref{tab:encoder-effect} shows the encoder effect on the standard scalar MSP and Energy decisions. On FGVC-Aircraft, replacing ResNet50 \cite{he2016resnet} with DINOv2 \cite{oquab2024dinov2} raises MSP known accuracy from 0.101 to 0.495, but also raises high-confidence false known acceptance rate (HC-FKAR) from 0.484 to 0.869. Stronger features make the classifier more competent and more confident, but confidence alone still does not certify that a predicted known class has enough evidence to accept a near-known unknown.

\begin{table}[t]
\centering
\small
\begin{tabular}{llrr}
\hline
Encoder & Method & Known Acc & HC-FKAR \\
\hline
ResNet50 & MSP & 0.101 & 0.484 \\
DINOv2 & MSP & 0.495 & 0.869 \\
ResNet50 & Energy & 0.204 & 0.928 \\
DINOv2 & Energy & 0.667 & 0.982 \\
\hline
\end{tabular}
\caption{Encoder effect on FGVC-Aircraft. Stronger DINOv2 features improve known accuracy but do not remove high-confidence near-known false acceptance.}
\label{tab:encoder-effect}
\end{table}

Recent post-hoc methods expose the same structural issue. Whether the score is MSP \cite{hendrycks2017baseline}, Energy \cite{liu2020energy}, activation-clipped confidence \cite{sun2021react}, ID-subspace residual \cite{wang2022vim}, neural-collapse alignment \cite{liu2025nci}, boundary distance \cite{liu2024fdbd}, or class-aware residual distance \cite{ling2025cadref}, the final decision is still a scalar score compared with a threshold. Such a decision can ask whether a sample looks globally known; it cannot ask whether the specific predicted known class has sufficient class-conditional evidence to accept it.

We propose EGUR-A, an Evidence-Guided Unknown Rejection framework. EGUR-A treats open-set recognition as known-class acceptance. Given a candidate class from a closed-set classifier, it verifies local class-conditional evidence and global residual evidence from the ID feature manifold. Because these two evidence sources have dataset-dependent reliability, EGUR-A selects their relative weight from known-sample evidence stability and coverage, without using unknown validation samples.

Our contributions are:
\begin{itemize}
\item We introduce HC-FKAR, a focused metric for high-confidence near-known unknown risk, and show that the risk persists across stronger encoders and modern scalar post-hoc methods.
\item We introduce EGUR-A, which separates closed-set confidence from known-class acceptance evidence and creates an explicit unsupported-known-like state for high-confidence samples lacking class-conditional support.
\item We show that local evidence and residual evidence have opposite reliability patterns across datasets, motivating an automatic evidence-weight selection rule rather than a fixed local-residual rule.
\item We show through controlled matched-KRR comparisons, naive fusion ablations, and operating-point curves that EGUR-A's gains are not reducible to a stricter threshold, scalar score averaging, or a single operating point.
\end{itemize}

\section{Related Work}

\paragraph{Scalar post-hoc detection.}
Post-hoc OOD and OSR methods often leave the backbone and classifier fixed, then reject samples with a scalar score. Classical scores include MSP \cite{hendrycks2017baseline}, Energy \cite{liu2020energy}, MaxLogit \cite{vaze2022openset}, prototype distance \cite{mensink2013distance}, and k-nearest-neighbor distance \cite{sun2022knn}. Recent variants improve the scalar through activation clipping \cite{sun2021react}, softmax entropy \cite{liu2023gen}, neighbor guidance \cite{park2023nnguide}, feature residuals \cite{wang2022vim}, decision-boundary cues \cite{liu2024fdbd}, activation scaling \cite{xu2024scale}, neural-collapse structure \cite{liu2025nci}, or class-aware relative features \cite{ling2025cadref}. These scores can be strong, especially residual scores on some datasets, but the final decision remains sample-level thresholding. EGUR-A instead verifies whether the predicted known class has local and global evidence to accept the sample.

\paragraph{Confidence calibration.}
Confidence calibration aims to align predicted confidence with empirical correctness \cite{guo2017calibration}, and recent surveys emphasize that class imbalance and hard samples can make calibration errors highly nonuniform \cite{dong2025calibrationsurvey}. This line of work is related because it also questions raw confidence as a decision signal. EGUR-A addresses a different failure mode: closed-set calibration can change the numerical meaning of a scalar confidence, but it does not by itself verify whether the predicted known class has sufficient open-set acceptance evidence. Our focus is therefore not better calibrated scoring, but replacing scalar acceptance with class-conditional evidence verification.

\paragraph{Training-time OSR.}
OpenMax \cite{bendale2016openmax}, ARPL \cite{chen2021arpl}, PROSER \cite{zhou2021proser}, and related methods change training objectives, synthesize unknown-like samples, or learn reciprocal structures. EGUR-A addresses a complementary setting: post-hoc unknown rejection on frozen features and an existing closed-set candidate classifier. We do not combine it with training-time OSR in the main experiments because the goal is to isolate the decision structure while holding representation learning fixed; combining the two is a natural downstream extension.

\paragraph{Semantic-shift evaluation.}
The Semantic Shift Benchmark (SSB) \cite{vaze2022openset} emphasizes unknown classes that are semantically close to known classes. Standard ranking metrics such as AUROC and FPR95 average over all unknowns. HC-FKAR instead focuses on the operationally risky subset: unknowns that the closed-set classifier itself treats as high-confidence known candidates.

\section{Method}

\subsection{From Scalar Knownness to Candidate-Specific Acceptance}

The central change in EGUR-A is the decision unit. Scalar methods make a sample-level knownness decision:
\begin{equation}
\operatorname{reject}(x) \Leftrightarrow s(x) \leq \tau,
\end{equation}
where $s(x)$ may be confidence, energy, residual, or another knownness score. EGUR-A instead makes a candidate-specific acceptance decision:
\begin{equation}
\operatorname{accept}(x,c) \Leftrightarrow c \text{ has sufficient evidence for } x,
\end{equation}
where $c=c(x)$ is the top-1 class proposed by the closed-set classifier. The question is no longer whether $x$ looks globally known; it is whether the predicted class $c$ has the right to accept $x$.

\begin{figure*}[t]
\centering
\includegraphics[width=0.98\textwidth]{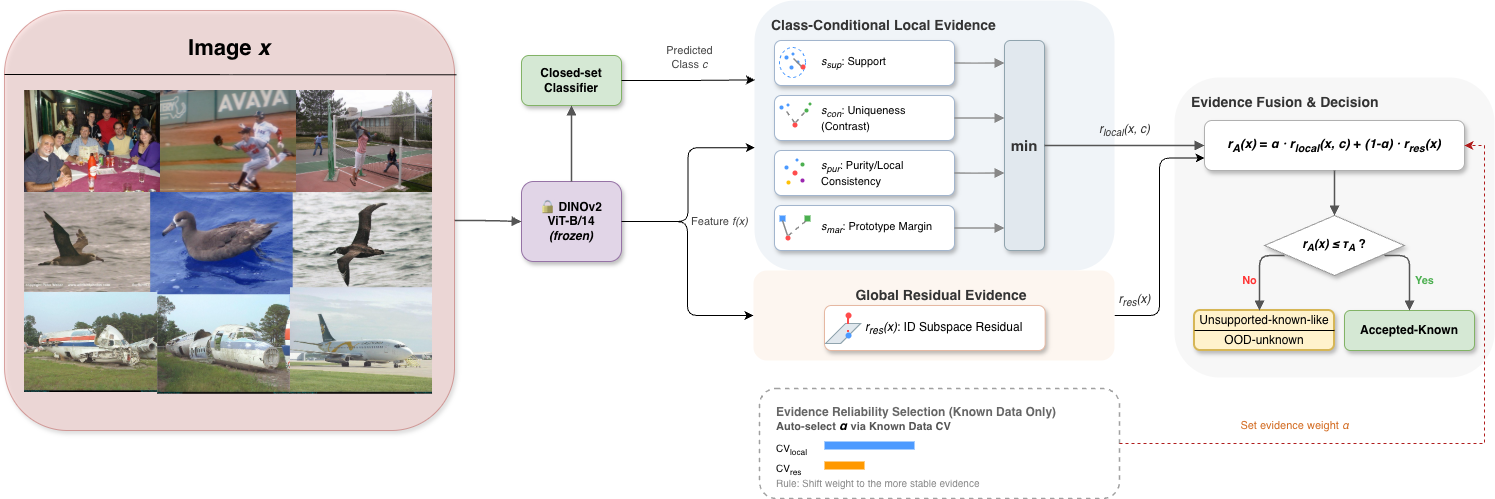}
\caption{EGUR-A architecture. A frozen DINOv2 encoder and closed-set classifier produce the feature and candidate class. EGUR-A computes class-conditional local evidence and global residual evidence, selects the evidence weight from known-data statistics, and accepts the sample only when the fused risk is below the calibrated threshold. The No branch groups two rejection states: unsupported-known-like for known-like samples without sufficient class-conditional evidence, and OOD-unknown for samples that do not remain known-like under the combined evidence test. The image panel is illustrative; inference is performed per test sample.}
\label{fig:egur-architecture}
\end{figure*}

Figure \ref{fig:egur-architecture} summarizes the EGUR-A inference pipeline. A frozen encoder extracts the feature, a closed-set classifier proposes the candidate class, class-conditional checks produce local risk, a global residual score produces residual risk, and an evidence weight selected from known data combines the two before the final acceptance decision.

The final branch in Figure \ref{fig:egur-architecture} still corresponds to three states. An \emph{accepted-known} sample is accepted by the candidate class. An \emph{unsupported-known-like} sample is high-confidence or known-like, but lacks sufficient class-conditional acceptance evidence. An \emph{out-of-distribution-unknown} sample lacks both known-like evidence and acceptance evidence. The middle state is the important addition: it names the failure mode that scalar thresholding collapses into either accepted or rejected.

\subsection{Necessary Conditions for Known-Class Acceptance}

Let $f(x)\in\mathbb{R}^d$ be a frozen feature and $\mathcal{D}^{\mathrm{tr}}_c$ the known training features of class $c$. Acceptance by class $c$ requires satisfying four necessary conditions. These conditions cover distinct ways in which candidate-class acceptance can fail: empirical coverage by the candidate class, exclusivity against nearby competitors, consistency of the local neighborhood, and class-level geometric separability. They are not treated as interchangeable scores. A severe failure of any active condition is sufficient to raise rejection risk.

\paragraph{Support.}
The candidate class must show that $x$ lies within its empirical support. We measure support by the candidate-class k-nearest-neighbor distance:
\begin{equation}
d_{\mathrm{sup}}(x,c)=\operatorname{kNNdist}(f(x),\mathcal{D}^{\mathrm{tr}}_c).
\end{equation}
If $x$ is outside the class-conditional support range, then class $c$ lacks local authority to accept it. Support thresholds are calibrated per class, so compact and diffuse classes are not forced onto the same distance scale.

\paragraph{Uniqueness.}
The candidate class must not merely be close; it must explain $x$ better than competing classes. EGUR-A compares candidate support to the nearest non-candidate support:
\begin{equation}
r_{\mathrm{con}}(x,c)=
\frac{d_{\mathrm{sup}}(x,c)}
{\min_{c'\neq c} d_{\mathrm{sup}}(x,c')}.
\end{equation}
A contrast ratio near 1 means that the candidate class and a competitor explain the sample similarly well, so the candidate has no unique acceptance claim.

\paragraph{Local consistency.}
The local neighborhood must support the top-1 prediction. We measure the fraction of candidate-class labels among the top-$m$ known neighbors:
\begin{equation}
p_{\mathrm{loc}}(x,c)=
\frac{1}{m}\sum_{z\in\mathcal{N}_m(x)}\mathbf{1}[y_z=c].
\end{equation}
Low purity indicates a mixed neighborhood, where classifier confidence is likely an unstable interpolation between nearby classes rather than a reliable class-membership signal.

\paragraph{Prototype advantage.}
The candidate class must also have class-level geometric advantage.
Let $\mu_c$ be the candidate centroid and let $c^-=\arg\min_{c'\neq c}d(f(x),\mu_{c'})$ denote the nearest competitor class. The prototype margin is
\begin{equation}
m_{\mathrm{proto}}(x,c)=
\frac{d(f(x),\mu_{c^-})-d(f(x),\mu_c)}
{d(f(x),\mu_{c^-})}.
\end{equation}
Small margin means the candidate and competitor are nearly equidistant, so the class-level decision is geometrically unstable.

These conditions are conjunction-like: acceptance requires all active conditions to hold. Let $[z]_0^1=\min(\max(z,0),1)$. EGUR-A converts each condition into a bounded evidence strength:
\begin{align}
s_{\mathrm{sup}} &= [1-d_{\mathrm{sup}}(x,c)/\tau_{\mathrm{sup}}(c)]_0^1, \\
s_{\mathrm{con}} &= [(\tau_{\mathrm{con}}-r_{\mathrm{con}}(x,c))/\tau_{\mathrm{con}}]_0^1, \\
s_{\mathrm{pur}} &= [(p_{\mathrm{loc}}(x,c)-\tau_{\mathrm{pur}})/(1-\tau_{\mathrm{pur}})]_0^1, \\
s_{\mathrm{mar}} &= [m_{\mathrm{proto}}(x,c)-\tau_{\mathrm{mar}}]_0^1 .
\end{align}
Disabled or inapplicable checks are omitted. The local risk is then
\begin{equation}
r_{\mathrm{local}}(x,c)=1-\min_j s_j .
\end{equation}
The minimum implements the acceptance semantics. A strong support score should not compensate for failed uniqueness, and a pure neighborhood should not compensate for being outside the candidate support. Mean aggregation would allow such compensation; minimum aggregation preserves the idea that any severe failure means class $c$ lacks sufficient right to accept $x$. It also gives the local chain complementary coverage: different conditions can expose different near-known acceptance failures, so a sample rejected by one condition need not be rejected by all others.

\subsection{Evidence Reliability Selection}

Local evidence asks whether the candidate class can accept $x$. It is still not sufficient by itself: a near-known unknown may pass local class-conditional checks while remaining globally inconsistent with the ID feature manifold. Global residual evidence supplies this complementary test. We fit an ID principal subspace on known training features and compute
\begin{equation}
\rho(x)=\|f(x)-\Pi_{\mathrm{ID}}f(x)\|_2.
\end{equation}
After normalization on known training statistics, this gives residual risk $r_{\mathrm{res}}(x)\in[0,1]$.

The two risks answer different questions and have complementary failure modes. Local evidence is unreliable when classes are diffuse and local neighborhoods mix across labels. Residual evidence is unreliable when unknown samples lie near the ID subspace despite lacking class-specific support. At matched KRR, residual-only gives HC-FKAR 0.338 on CUB but 0.070 on Aircraft; a fixed local-residual AND combination reverses this pattern to 0.186 and 0.189, respectively. Neither ``always trust local'' nor ``always trust residual'' is reliable.

We call $\alpha$ the \emph{evidence weight}: it controls how much the acceptance decision relies on local class-conditional evidence versus global residual evidence. EGUR-A combines the two risks as
\begin{equation}
r_A(x)=\alpha r_{\mathrm{local}}(x,c(x))+(1-\alpha)r_{\mathrm{res}}(x),
\end{equation}
and accepts when
\begin{equation}
r_A(x)\leq \tau_A.
\end{equation}
The threshold $\tau_A$ is calibrated on known data. No unknown validation samples are used.

The dataset-level evidence weight is selected automatically from known samples. Let $G=\{0.2,0.4,0.6,0.8,1.0\}$ be the evidence-weight grid. First, we compare evidence stability on the known calibration samples using coefficients of variation:
\begin{equation}
\mathrm{CV}_{\mathrm{local}}=
\frac{\operatorname{std}(r_{\mathrm{local}})}{\operatorname{mean}(r_{\mathrm{local}})},\quad
\mathrm{CV}_{\mathrm{res}}=
\frac{\operatorname{std}(r_{\mathrm{res}})}{\operatorname{mean}(r_{\mathrm{res}})} .
\end{equation}
If residual risk is more stable on known samples, $\mathrm{CV}_{\mathrm{res}}<\mathrm{CV}_{\mathrm{local}}$, we select the residual-dominant endpoint $\alpha=0.2$. Otherwise, we compute the known accuracy obtained by each $\alpha\in G$ after calibrating $\tau_A$ to the target KRR on known samples, choose the known-accuracy-maximizing $\alpha_{\mathrm{KA}}$, and move one grid step toward residual evidence:
\begin{equation}
\alpha=\max(\alpha_{\mathrm{KA}}-0.2,0.2).
\end{equation}
The one-step correction is fixed. It counteracts the empirical bias of known-only selection, which otherwise favors local evidence weight that improves known coverage but can degrade high-confidence unknown rejection; for example, on Aircraft, known-only selection would choose $\alpha=0.8$ and raise HC-FKAR from 0.071 to 0.110 (see the supporting material). This selection procedure uses only known labels and cached risk scores; fixed evidence-weight sweeps are used only as diagnostics.

Training-set compactness remains an interpretable correlate of the selected regimes, but it is not used as a fitted cross-dataset formula. In compact class structures, local evidence tends to receive higher weight; in diffuse structures, residual evidence becomes more reliable. The rule selects $\alpha=0.8,0.6,0.2$ on CUB, Aircraft, and ImageNet-hard, respectively, and also selects $\alpha=0.8$ on held-out Stanford Cars; the per-dataset CV statistics are reported in the supporting material. ImageNet-hard illustrates the residual-stable case: the automatic rule places 80\% of the evidence weight on global residual risk, so EGUR-A approaches a residual verifier by design.

\section{Experiments}

\begin{table*}[!t]
\centering
\small
\begin{tabular}{lrrrrrr}
\hline
Method & \shortstack{CUB\\KRR} & \shortstack{CUB\\HC-FKAR} & \shortstack{Aircraft\\KRR} & \shortstack{Aircraft\\HC-FKAR} & \shortstack{ImageNet-hard\\KRR} & \shortstack{ImageNet-hard\\HC-FKAR} \\
\hline
MSP & 0.238 & 0.442 & 0.475 & 0.869 & 0.263 & 0.763 \\
Energy & 0.145 & 0.622 & 0.221 & 0.982 & 0.164 & 0.577 \\
NNGuide & 0.117 & 0.695 & 0.197 & 0.981 & 0.139 & 0.604 \\
NCI-OOD & 0.056 & 0.963 & 0.058 & 0.951 & 0.055 & 0.980 \\
CADRef & 0.077 & 0.800 & 0.116 & 0.988 & 0.100 & 0.668 \\
ViM & 0.391 & 0.258 & 0.837 & 0.029 & 0.130 & 0.525 \\
EGUR-A & 0.411 & \textbf{0.198} & 0.698 & \textbf{0.071} & 0.238 & \textbf{0.336} \\
\hline
\end{tabular}
\caption{Default-workpoint stress test. Baselines use natural/default thresholds; EGUR-A uses its calibrated operating point. This is not a matched-KRR comparison: it shows that representative scalar methods repeatedly accept high-confidence near-known unknowns. ViM's Aircraft HC-FKAR 0.029 occurs only at KRR 0.837 and Known Acc 0.150. Matched-KRR comparisons are in Tables \ref{tab:main-cub}--\ref{tab:main-imagenet}; full records are in the supporting material.}
\label{tab:default-baselines}
\end{table*}

\subsection{Setup and Metrics}

\paragraph{Backbone and datasets.}
All main experiments use frozen DINOv2 ViT-B/14-reg4 features \cite{oquab2024dinov2,darcet2024registers} and a linear probe as the closed-set candidate generator. We evaluate on CUB SSB \cite{wah2011cub,vaze2022openset} (100 known / 100 unknown classes), FGVC-Aircraft SSB \cite{maji2013aircraft,vaze2022openset} (50/50), and ImageNet SSB-hard compact \cite{deng2009imagenet,vaze2022openset}. ImageNet-hard uses 500 known classes and 49,000 hard unknown samples, with 50 samples per hard unknown class from the official SSB hard split. DTD \cite{cimpoi2014dtd} is used as a far-OOD safety check. Stanford Cars \cite{krause2013cars} is used only for the held-out evidence-weight rule check.

\paragraph{Baselines.}
We compare against scalar post-hoc scores including MSP, Energy, MaxLogit \cite{vaze2022openset}, prototype distance \cite{mensink2013distance}, k-nearest-neighbor distance \cite{sun2022knn}, ReAct \cite{sun2021react}, GEN \cite{liu2023gen}, NNGuide \cite{park2023nnguide}, ViM \cite{wang2022vim}, fDBD \cite{liu2024fdbd}, SCALE \cite{xu2024scale}, NCI-OOD \cite{liu2025nci}, and CADRef \cite{ling2025cadref}. Table \ref{tab:default-baselines} reports the main default-workpoint stress-test baselines. Auxiliary classical distance baselines, aggregate metrics, Pareto points, and negative-result records are provided in the supporting material.

\paragraph{Metrics.}
Known Acc is the fraction of known samples correctly classified and accepted. KRR is known rejection rate. FKAR is false known acceptance rate over all unknowns. Let $\mathcal{U}$ be the unknown set, $a(x)\in\{0,1\}$ indicate acceptance as known, and $q(x)$ be closed-set confidence. We define
\begin{equation}
\mathrm{HC\mbox{-}FKAR}@t=
\frac{|\{x\in\mathcal{U}:q(x)\geq t\wedge a(x)=1\}|}
{|\{x\in\mathcal{U}:q(x)\geq t\}|}.
\end{equation}
In words, HC-FKAR@$t$ is the fraction of unknown samples whose closed-set confidence meets or exceeds $t$ and are still accepted as known. The denominator is the high-confidence unknown subset at threshold $t$; if no unknown reaches $t$, the metric is undefined.
AUROC and FPR95 summarize global ranking over all unknowns, but they do not isolate the high-confidence subpopulation where near-known errors are most operationally risky. Aggregate metrics are reported in the supporting material as a sanity check: on CUB, EGUR-A's FPR95 is 0.491, lower than MSP (0.824) and ViM (0.615), while AUROC differences reflect that EGUR-A does not optimize over the full unknown distribution. We therefore use HC-FKAR at matched or comparable KRR as the primary comparison.

The main tables report HC-FKAR@0.90. Figure \ref{fig:hc-threshold-sensitivity} reports HC-FKAR@$t$ for $t\in\{0.80,0.85,0.90,0.95,0.99\}$; EGUR-A remains below MSP and ViM across this range on all three datasets.

\subsection{Scalar Thresholds Fail at High Confidence}

Table \ref{tab:default-baselines} evaluates default or natural operating points. The point is not that one baseline is careless; the point is that scalar thresholding repeatedly accepts high-confidence near-known unknowns. On Aircraft, representative scalar methods such as Energy, NNGuide, NCI-OOD, and CADRef all exceed 0.95 HC-FKAR at their natural operating points; the full supporting table shows the same pattern for additional scalar baselines such as fDBD and SCALE. ViM has a low default Aircraft HC-FKAR, but only at KRR 0.837 and Known Acc 0.150, an impractical acceptance regime. This table establishes the scope of the structural blind spot.

\subsection{Main Matched-KRR Results}

Tables \ref{tab:main-cub}, \ref{tab:main-aircraft}, and \ref{tab:main-imagenet} compare EGUR-A with all available linear-probe post-hoc baselines at matched known rejection rates. The EGUR-A rows use the same calibrated operating points as in Table \ref{tab:default-baselines}; all scalar baselines are re-thresholded to the corresponding EGUR-A KRR. These methods share the same closed-set linear-probe confidence, so HC-FKAR is measured on the same high-confidence subset. EGUR-A achieves the lowest HC-FKAR on all three datasets. ViM is the strongest residual scalar baseline; because EGUR-A also includes residual evidence, gains over ViM isolate class-conditional acceptance evidence. The margin is largest on Aircraft, where EGUR-A reduces HC-FKAR to 0.071. On ImageNet-hard, stratified bootstrap over unknown classes gives a stable EGUR-A advantage over ViM; details are in the supporting material.

\begin{table}[t]
\centering
\scriptsize
\setlength{\tabcolsep}{3pt}
\begin{tabular}{lrrrr}
\hline
Method & Known Acc & KRR & FKAR & HC-FKAR \\
\hline
MSP & 0.586 & 0.411 & \textbf{0.097} & 0.261 \\
Energy & 0.582 & 0.411 & 0.147 & 0.333 \\
MaxLogit & 0.582 & 0.411 & 0.137 & 0.331 \\
GEN & 0.583 & 0.411 & 0.139 & 0.331 \\
ReAct-Energy & 0.581 & 0.411 & 0.148 & 0.335 \\
NNGuide & 0.578 & 0.411 & 0.189 & 0.367 \\
ViM & 0.580 & 0.411 & 0.129 & 0.246 \\
fDBD & 0.586 & 0.411 & 0.120 & 0.284 \\
SCALE-Energy & 0.576 & 0.411 & 0.180 & 0.378 \\
NCI-OOD & 0.558 & 0.411 & 0.666 & 0.726 \\
CADRef & 0.577 & 0.411 & 0.161 & 0.336 \\
EGUR-A, $\alpha=0.8$ & 0.580 & 0.411 & 0.116 & \textbf{0.198} \\
\hline
\end{tabular}
\caption{CUB matched-KRR results. All baselines are re-thresholded to EGUR-A's KRR.}
\label{tab:main-cub}
\end{table}

\begin{table}[t]
\centering
\scriptsize
\setlength{\tabcolsep}{3pt}
\begin{tabular}{lrrrr}
\hline
Method & Known Acc & KRR & FKAR & HC-FKAR \\
\hline
MSP & 0.295 & 0.698 & 0.116 & 0.326 \\
Energy & 0.282 & 0.698 & 0.185 & 0.466 \\
MaxLogit & 0.283 & 0.698 & 0.183 & 0.472 \\
GEN & 0.286 & 0.698 & 0.160 & 0.421 \\
ReAct-Energy & 0.282 & 0.698 & 0.187 & 0.470 \\
NNGuide & 0.280 & 0.698 & 0.209 & 0.482 \\
ViM & 0.274 & 0.698 & \textbf{0.078} & 0.097 \\
fDBD & 0.301 & 0.698 & 0.245 & 0.570 \\
SCALE-Energy & 0.277 & 0.698 & 0.213 & 0.518 \\
NCI-OOD & 0.237 & 0.698 & 0.237 & 0.352 \\
CADRef & 0.281 & 0.698 & 0.210 & 0.364 \\
EGUR-A, $\alpha=0.6$ & 0.270 & 0.698 & 0.088 & \textbf{0.071} \\
\hline
\end{tabular}
\caption{FGVC-Aircraft matched-KRR results. EGUR-A achieves the lowest HC-FKAR; the FKAR difference from ViM reflects its high-confidence focus.}
\label{tab:main-aircraft}
\end{table}

\begin{table}[t]
\centering
\scriptsize
\setlength{\tabcolsep}{3pt}
\begin{tabular}{lrrrr}
\hline
Method & Known Acc & KRR & FKAR & HC-FKAR \\
\hline
MSP & 0.703 & 0.238 & 0.241 & 0.847 \\
Energy & 0.688 & 0.238 & 0.139 & 0.441 \\
MaxLogit & 0.689 & 0.238 & 0.139 & 0.447 \\
GEN & 0.690 & 0.238 & 0.146 & 0.468 \\
ReAct-Energy & 0.688 & 0.238 & 0.139 & 0.440 \\
NNGuide & 0.688 & 0.238 & 0.136 & 0.426 \\
ViM & 0.685 & 0.238 & \textbf{0.116} & 0.360 \\
fDBD & 0.694 & 0.238 & 0.151 & 0.476 \\
SCALE-Energy & 0.685 & 0.238 & 0.147 & 0.460 \\
NCI-OOD & 0.646 & 0.238 & 0.880 & 0.868 \\
CADRef & 0.692 & 0.238 & 0.134 & 0.419 \\
EGUR-A, $\alpha=0.2$$^\dagger$ & 0.668 & 0.238 & 0.127 & \textbf{0.336} \\
\hline
\end{tabular}
\caption{ImageNet-hard matched-KRR results. EGUR-A has the lowest HC-FKAR at $\alpha=0.2$. MSP is re-thresholded below its natural KRR, which raises its HC-FKAR relative to Table \ref{tab:default-baselines}. $^\dagger$Bootstrap details are in the supporting material.}
\label{tab:main-imagenet}
\end{table}

\subsection{High-Confidence Stress Test}

\begin{figure*}[t]
\centering
\includegraphics[width=0.92\textwidth]{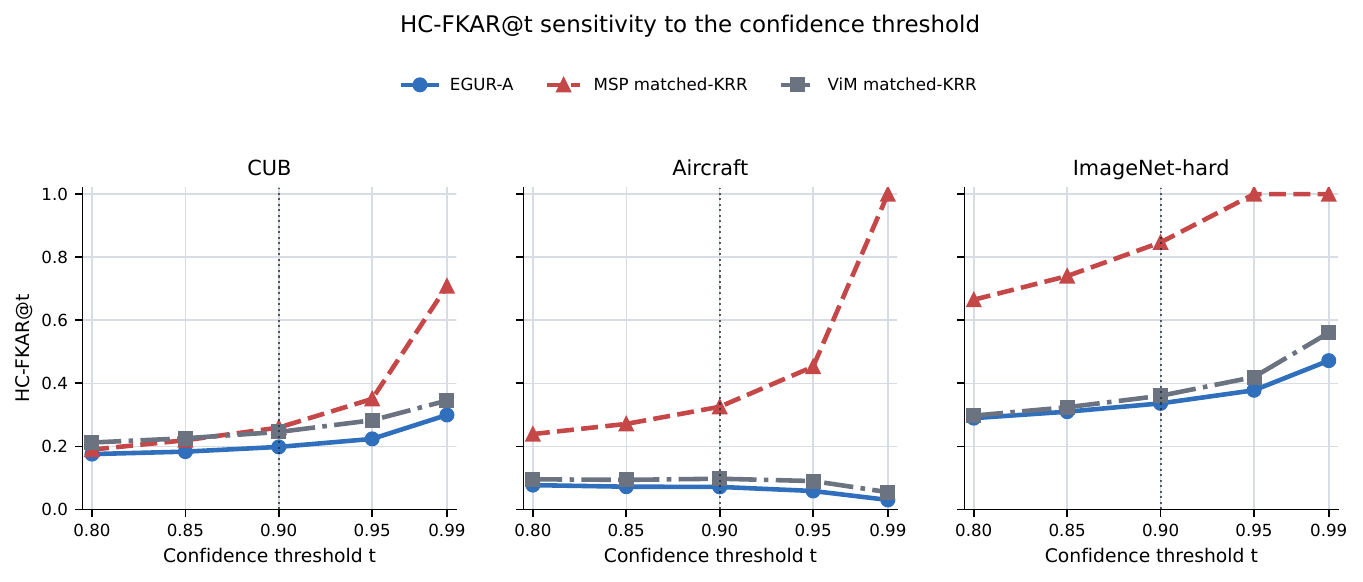}
\caption{HC-FKAR@$t$ sensitivity across confidence thresholds. Curves report cumulative false known acceptance among unknown samples with closed-set confidence at least $t$. MSP is the standard scalar reference; ViM is the strongest matched-KRR scalar baseline. EGUR-A remains below both across all thresholds and datasets, including the extreme $t=0.99$ tail. The dotted line marks the main reported threshold $t=0.90$.}
\label{fig:hc-threshold-sensitivity}
\end{figure*}

EGUR-A's effect is concentrated exactly where scalar confidence is most misleading. At matched KRR, EGUR-A rejects the majority of MSP-accepted high-confidence unknowns as unsupported-known-like: 375 of 388 on Aircraft and 189 of 281 on CUB. Figure \ref{fig:hc-threshold-sensitivity} shows the cumulative HC-FKAR@$t$ curve. EGUR-A remains below MSP and ViM from $t=0.80$ to $0.99$ on all three datasets; at the extreme $t=0.99$ tail, MSP reaches 0.708/1.000/1.000 on CUB/Aircraft/ImageNet-hard, while EGUR-A reduces these values to 0.299/0.030/0.472. Complete confidence-bin tables are in the supporting material.

\subsection{Operating-Point Evidence}

Known accuracy and HC-FKAR trade off through the acceptance threshold. Operating-point curves sweeping target known-accuracy ranges are provided in the supporting material and confirm that EGUR-A's advantage over residual-only persists across practical ranges on CUB and Aircraft, while ImageNet-hard stays close to the residual endpoint expected from $\alpha=0.2$.

\subsection{Ablations}

\paragraph{Evidence Structure vs. Score Averaging.}

EGUR-A is not simply a weighted sum of MSP confidence and residual knownness. A naive fusion of normalized MSP confidence and residual knownness gives matched-KRR HC-FKAR of 0.233, 0.207, and 0.323 on CUB, Aircraft, and ImageNet-hard, compared with EGUR-A's 0.198, 0.071, and 0.336. The large Aircraft gap shows that EGUR-A is not reducible to scalar score averaging, while ImageNet-hard again behaves as a residual-saturation regime.

\paragraph{Local Evidence Modules.}

Aircraft local-evidence ablations show that support is necessary but insufficient: removing support raises HC-FKAR to 0.619, while support alone gives 0.537. Adding contrast, purity, and margin reduces HC-FKAR to 0.241 in EGUR-min; removing contrast and purity gives 0.320, removing margin gives 0.289, and global calibration gives 0.577. The final drop to 0.071 comes from adaptive residual weighting in EGUR-A.

\paragraph{Evidence-Weight Rule Check and Far-OOD Safety.}

The automatic evidence-weight rule uses only known-sample statistics. Fixed evidence-weight sweeps are diagnostic rather than part of selection: the rule selects the HC-FKAR-optimal fixed evidence weight on Aircraft, ImageNet-hard, and held-out Stanford Cars, and selects $\alpha=0.8$ instead of $\alpha=1.0$ on CUB, increasing HC-FKAR by only 0.002 (0.196 to 0.198). The corresponding CUB overall FKAR increases from 0.088 to 0.116, but remains below both MSP (0.163) and ViM (0.138). This small cost removes the need for cross-dataset calibration constants.

EGUR-A also preserves far-OOD rejection in the DTD check. CUB EGUR-A yields DTD-FKAR 0.000, and Aircraft EGUR-A with $\alpha=0.6$ also yields DTD-FKAR 0.000. The class-conditional acceptance mechanism therefore does not introduce far-OOD acceptance regression in this setting.

\section{Discussion}

\paragraph{Scalar saturation and metric scope.}
The default-workpoint results show that the high-confidence failure is not a defect of a single baseline. fDBD, SCALE, NCI-OOD, and CADRef all exceed 0.95 HC-FKAR on Aircraft (see supporting material), and MSP saturates to FKAR 1.000 in the highest-confidence Aircraft and ImageNet-hard bins. These methods use different scalar signals, but share the same final structure: accept when one global score clears a threshold. EGUR-A instead asks whether the predicted class has acceptance evidence. This does not imply universal dominance on every aggregate metric. On Aircraft and ImageNet-hard, EGUR-A has lower HC-FKAR than the residual scalar baseline at matched KRR (0.071 vs. 0.097; 0.336 vs. 0.360), but slightly higher overall FKAR (0.088 vs. 0.078; 0.127 vs. 0.116), reflecting its focus on high-confidence known-like errors.

\paragraph{Complementary coverage and residual saturation.}
Local evidence helps through complementary coverage: under weakest-evidence aggregation, each condition can expose an acceptance failure that the others miss. This mechanism is most useful when known-class supports have sufficiently distinct local geometry. CUB and Aircraft show this behavior in different ways: CUB is local-dominant, while Aircraft benefits from combining local checks with residual evidence. ImageNet-hard is different. With 500 known classes and heavily overlapping supports inside the ID manifold, local complementary coverage becomes less informative, and the automatic evidence-weight rule selects $\alpha=0.2$, making EGUR-A residual-dominant by design. The naive pure-residual point reaches HC-FKAR 0.323, only 0.013 below EGUR-A's 0.336. We interpret this as a residual-saturation endpoint, not as a contradiction: removing class-conditional evidence can slightly improve this particular residual-dominated benchmark, but it discards the acceptance structure that matters in local-dominant regimes. In the Evidence Reliability Selection diagnostic, pure residual thresholding gives CUB HC-FKAR 0.338, far above EGUR-A's 0.198; this local-dominant cost is much larger than the 0.013 ImageNet-hard gap. EGUR-A therefore exposes a regime diagnosis: when local geometry supports complementary coverage, class-conditional evidence is valuable; when residual evidence saturates, the rule shifts toward the residual endpoint while preserving a unified acceptance framework.

\paragraph{Known accuracy and rejection.}
Known Acc drops because EGUR-A rejects more known samples; relative to MSP, the drops are close to the corresponding KRR increases. At matched or comparable KRR, EGUR-A's Known Acc is within about 0.02 of the residual scalar baseline on all three datasets, indicating that the acceptance structure does not introduce substantial additional known-sample loss beyond the rejection rate itself. A deployment prioritizing coverage can relax $\tau_A$, while a deployment prioritizing high-confidence safety can use the reported operating points.

\paragraph{Unknown-free evidence-weight selection.}
The automatic evidence-weight rule replaces cross-dataset fitting with within-dataset known-sample statistics. It should be read as a practical approximation rather than a theoretical characterization of the local-residual reliability boundary. The fundamental difficulty is that the optimal evidence weight depends on the near-known unknown distribution, which is unavailable at deployment; any known-sample proxy for that distribution introduces an unverifiable gap. We therefore make the selection rule reproducible and quantify its cost. Fixed evidence-weight sweeps show that the rule selects the HC-FKAR-optimal evidence weight on Aircraft, ImageNet-hard, and held-out Cars, and differs by only 0.002 on CUB, where it selects $\alpha=0.8$ instead of the oracle $\alpha=1.0$.

\paragraph{Limitations.}
EGUR-A is a post-hoc acceptance verifier on frozen features, so it cannot separate unknowns that are fully overlapped with known classes in the feature space. Two additional boundaries remain. First, evidence-weight selection is unknown-free but not theoretically optimal: without observing the near-known unknown distribution, no known-sample diagnostic can exactly locate the local-residual reliability boundary. Second, the effective domain of hard-min complementary coverage depends on the local geometry of known-class support boundaries. A full theory would need to characterize which directions of near-known incursion cause gate-score reversals at those boundaries, likely requiring tools from local manifold geometry beyond the empirical diagnostics used here. Training-time use of near-known hard negatives is a complementary direction for improving representation geometry before post-hoc acceptance.

\section{Conclusion}

High-confidence near-known unknowns reveal a mismatch at the heart of scalar-threshold OSR: high classifier confidence does not imply that the predicted known class has sufficient acceptance evidence. EGUR-A separates these two notions by verifying class-conditional local evidence and global residual evidence under an unknown-free evidence-weight selection rule. Across fine-grained and large-scale semantic-shift benchmarks, this decision structure substantially reduces HC-FKAR at matched known-rejection operating points, while making the local-dominant and residual-saturation regimes explicit. The result is not just a stronger threshold; it is a clearer way to ask whether a known class is entitled to accept a sample.

\bibliography{aaai2026}

@inproceedings{hendrycks2017baseline,
  author    = {Hendrycks, Dan and Gimpel, Kevin},
  title     = {A Baseline for Detecting Misclassified and Out-of-Distribution Examples in Neural Networks},
  booktitle = {International Conference on Learning Representations},
  year      = {2017}
}

@inproceedings{liu2020energy,
  author    = {Liu, Weitang and Wang, Xiaoyun and Owens, John D. and Li, Yixuan},
  title     = {Energy-Based Out-of-Distribution Detection},
  booktitle = {Advances in Neural Information Processing Systems},
  volume    = {33},
  pages     = {21464--21475},
  year      = {2020}
}

@inproceedings{bendale2016openmax,
  author    = {Bendale, Abhijit and Boult, Terrance E.},
  title     = {Towards Open Set Deep Networks},
  booktitle = {Proceedings of the IEEE Conference on Computer Vision and Pattern Recognition},
  pages     = {1563--1572},
  year      = {2016}
}

@inproceedings{wang2022vim,
  author    = {Wang, Haoqi and Li, Zhizhong and Feng, Litong and Zhang, Wayne},
  title     = {{ViM}: Out-of-Distribution with Virtual-Logit Matching},
  booktitle = {Proceedings of the IEEE/CVF Conference on Computer Vision and Pattern Recognition},
  pages     = {4921--4930},
  year      = {2022}
}

@inproceedings{sun2021react,
  author    = {Sun, Yiyou and Guo, Chuan and Li, Yixuan},
  title     = {{ReAct}: Out-of-Distribution Detection with Rectified Activations},
  booktitle = {Advances in Neural Information Processing Systems},
  volume    = {34},
  pages     = {144--157},
  year      = {2021}
}

@inproceedings{liu2023gen,
  author    = {Liu, Xixi and Lochman, Yaroslava and Zach, Christopher},
  title     = {{GEN}: Pushing the Limits of Softmax-Based Out-of-Distribution Detection},
  booktitle = {Proceedings of the IEEE/CVF Conference on Computer Vision and Pattern Recognition},
  pages     = {23946--23955},
  year      = {2023}
}

@inproceedings{park2023nnguide,
  author    = {Park, Jaewoo and Jung, Yoon Gyo and Teoh, Andrew Beng Jin},
  title     = {Nearest Neighbor Guidance for Out-of-Distribution Detection},
  booktitle = {Proceedings of the IEEE/CVF International Conference on Computer Vision},
  pages     = {1686--1695},
  year      = {2023}
}

@inproceedings{liu2024fdbd,
  author    = {Liu, Litian and Qin, Yao},
  title     = {Fast Decision Boundary Based Out-of-Distribution Detector},
  booktitle = {Proceedings of the 41st International Conference on Machine Learning},
  series    = {Proceedings of Machine Learning Research},
  volume    = {235},
  pages     = {31728--31746},
  publisher = {PMLR},
  year      = {2024}
}

@inproceedings{xu2024scale,
  author    = {Xu, Kai and Chen, Rongyu and Franchi, Gianni and Yao, Angela},
  title     = {{SCALE}: Scaling for Training Time and Post-hoc Out-of-distribution Detection Enhancement},
  booktitle = {International Conference on Learning Representations},
  year      = {2024}
}

@inproceedings{liu2025nci,
  author    = {Liu, Litian and Qin, Yao},
  title     = {Detecting Out-of-Distribution Through the Lens of Neural Collapse},
  booktitle = {Proceedings of the IEEE/CVF Conference on Computer Vision and Pattern Recognition},
  pages     = {15424--15433},
  year      = {2025}
}

@inproceedings{ling2025cadref,
  author    = {Ling, Zhiwei and Chang, Yachen and Zhao, Hailiang and Zhao, Xinkui and Chow, Kingsum and Deng, Shuiguang},
  title     = {{CADRef}: Robust Out-of-Distribution Detection via Class-Aware Decoupled Relative Feature Leveraging},
  booktitle = {Proceedings of the IEEE/CVF Conference on Computer Vision and Pattern Recognition},
  pages     = {4968--4977},
  year      = {2025}
}

@article{chen2021arpl,
  author    = {Chen, Guangyao and Peng, Peixi and Wang, Xiangqian and Tian, Yonghong},
  title     = {Adversarial Reciprocal Points Learning for Open Set Recognition},
  journal   = {IEEE Transactions on Pattern Analysis and Machine Intelligence},
  volume    = {44},
  number    = {11},
  pages     = {8065--8081},
  year      = {2022},
  doi       = {10.1109/TPAMI.2021.3106743}
}

@inproceedings{zhou2021proser,
  author    = {Zhou, Da-Wei and Ye, Han-Jia and Zhan, De-Chuan},
  title     = {Learning Placeholders for Open-Set Recognition},
  booktitle = {Proceedings of the IEEE/CVF Conference on Computer Vision and Pattern Recognition},
  pages     = {4401--4410},
  year      = {2021}
}

@inproceedings{vaze2022openset,
  author    = {Vaze, Sagar and Han, Kai and Vedaldi, Andrea and Zisserman, Andrew},
  title     = {Open-Set Recognition: A Good Closed-Set Classifier Is All You Need},
  booktitle = {International Conference on Learning Representations},
  year      = {2022}
}

@article{oquab2024dinov2,
  author        = {Oquab, Maxime and Darcet, Timoth{\'e}e and Moutakanni, Th{\'e}o and Vo, Huy and Szafraniec, Marc and Khalidov, Vasil and Fernandez, Pierre and Haziza, Daniel and Massa, Francisco and El-Nouby, Alaaeldin and Assran, Mahmoud and Ballas, Nicolas and Galuba, Wojciech and Howes, Russell and Huang, Po-Yao and Li, Shang-Wen and Misra, Ishan and Rabbat, Michael and Sharma, Vasu and Synnaeve, Gabriel and Xu, Hu and Jegou, Herve and Mairal, Julien and Labatut, Patrick and Joulin, Armand and Bojanowski, Piotr},
  title         = {{DINOv2}: Learning Robust Visual Features without Supervision},
  journal       = {Transactions on Machine Learning Research},
  year          = {2024},
  eprint        = {2304.07193},
  archivePrefix = {arXiv},
  primaryClass  = {cs.CV},
  url           = {https://openreview.net/forum?id=a68SUt6zFt}
}

@inproceedings{darcet2024registers,
  author    = {Darcet, Timoth{\'e}e and Oquab, Maxime and Mairal, Julien and Bojanowski, Piotr},
  title     = {Vision Transformers Need Registers},
  booktitle = {International Conference on Learning Representations},
  year      = {2024},
  url       = {https://openreview.net/forum?id=2dnO3LLiJ1}
}

@inproceedings{he2016resnet,
  author    = {He, Kaiming and Zhang, Xiangyu and Ren, Shaoqing and Sun, Jian},
  title     = {Deep Residual Learning for Image Recognition},
  booktitle = {Proceedings of the IEEE Conference on Computer Vision and Pattern Recognition},
  pages     = {770--778},
  year      = {2016},
  doi       = {10.1109/CVPR.2016.90}
}

@techreport{wah2011cub,
  author    = {Wah, Catherine and Branson, Steve and Welinder, Peter and Perona, Pietro and Belongie, Serge},
  title     = {The Caltech-UCSD Birds-200-2011 Dataset},
  institution = {California Institute of Technology},
  number    = {CNS-TR-2011-001},
  year      = {2011}
}

@article{maji2013aircraft,
  author    = {Maji, Subhransu and Rahtu, Esa and Kannala, Juho and Blaschko, Matthew and Vedaldi, Andrea},
  title     = {Fine-Grained Visual Classification of Aircraft},
  journal   = {arXiv preprint arXiv:1306.5151},
  year      = {2013},
  eprint    = {1306.5151},
  archivePrefix = {arXiv},
  primaryClass = {cs.CV}
}

@inproceedings{krause2013cars,
  author    = {Krause, Jonathan and Stark, Michael and Deng, Jia and Fei-Fei, Li},
  title     = {3D Object Representations for Fine-Grained Categorization},
  booktitle = {IEEE International Conference on Computer Vision Workshops},
  pages     = {554--561},
  doi       = {10.1109/ICCVW.2013.77},
  year      = {2013}
}

@inproceedings{cimpoi2014dtd,
  author    = {Cimpoi, Mircea and Maji, Subhransu and Kokkinos, Iasonas and Mohamed, Sammy and Vedaldi, Andrea},
  title     = {Describing Textures in the Wild},
  booktitle = {Proceedings of the IEEE Conference on Computer Vision and Pattern Recognition},
  pages     = {3606--3613},
  year      = {2014}
}

@inproceedings{deng2009imagenet,
  author    = {Deng, Jia and Dong, Wei and Socher, Richard and Li, Li-Jia and Li, Kai and Fei-Fei, Li},
  title     = {ImageNet: A Large-Scale Hierarchical Image Database},
  booktitle = {Proceedings of the IEEE Conference on Computer Vision and Pattern Recognition},
  pages     = {248--255},
  doi       = {10.1109/CVPR.2009.5206848},
  year      = {2009}
}

@inproceedings{sun2022knn,
  author    = {Sun, Yiyou and Ming, Yifei and Zhu, Xiaojin and Li, Yixuan},
  title     = {Out-of-Distribution Detection with Deep Nearest Neighbors},
  booktitle = {Proceedings of the 39th International Conference on Machine Learning},
  series    = {Proceedings of Machine Learning Research},
  volume    = {162},
  pages     = {20827--20840},
  publisher = {PMLR},
  year      = {2022}
}

@article{mensink2013distance,
  author  = {Mensink, Thomas and Verbeek, Jakob and Perronnin, Florent and Csurka, Gabriela},
  title   = {Distance-Based Image Classification: Generalizing to New Classes at Near-Zero Cost},
  journal = {IEEE Transactions on Pattern Analysis and Machine Intelligence},
  volume  = {35},
  number  = {11},
  pages   = {2624--2637},
  year    = {2013},
  doi     = {10.1109/TPAMI.2013.83}
}

@inproceedings{guo2017calibration,
  author    = {Guo, Chuan and Pleiss, Geoff and Sun, Yu and Weinberger, Kilian Q.},
  title     = {On Calibration of Modern Neural Networks},
  booktitle = {Proceedings of the 34th International Conference on Machine Learning},
  series    = {Proceedings of Machine Learning Research},
  volume    = {70},
  pages     = {1321--1330},
  publisher = {PMLR},
  year      = {2017}
}

@article{dong2025calibrationsurvey,
  author  = {Dong, Jinzong and Jiang, Zhaohui and Pan, Dong and Chen, Zhiwen and Guan, Qingyi and Zhang, Hongbin and Gui, Gui and Gui, Weihua},
  title   = {A Survey on Confidence Calibration of Deep Learning-Based Classification Models Under Class Imbalance Data},
  journal = {IEEE Transactions on Neural Networks and Learning Systems},
  volume  = {36},
  number  = {9},
  pages   = {15664--15684},
  year    = {2025},
  doi     = {10.1109/TNNLS.2025.3565159}
}

\end{document}


\maketitle

\section{Purpose}

This supplement records experiment details and auxiliary tables that support the main paper. The primary paper focuses on HC-FKAR at matched or comparable KRR; this document keeps the broader default-workpoint, aggregate-metric, evidence-weight-selection, and negative-result records visible.

\section{Local Risk Implementation}

The main paper states the local evidence checks and gives the risk aggregation rule. For implementation clarity, the active evidence strengths are:
\begin{align}
s_{\mathrm{sup}} &= [1-d_{\mathrm{sup}}(x,c)/\tau_{\mathrm{sup}}(c)]_0^1, \\
s_{\mathrm{con}} &= [(\tau_{\mathrm{con}}-r_{\mathrm{con}}(x,c))/\tau_{\mathrm{con}}]_0^1, \\
s_{\mathrm{pur}} &= [(p_{\mathrm{loc}}(x,c)-\tau_{\mathrm{pur}})/(1-\tau_{\mathrm{pur}})]_0^1, \\
s_{\mathrm{mar}} &= [m_{\mathrm{proto}}(x,c)-\tau_{\mathrm{mar}}]_0^1 .
\end{align}
When the optional local-conflict check is active, we also use
\begin{equation}
s_{\mathrm{conf}}=[(\tau_{\mathrm{conf}}-\ell_{\mathrm{conf}}(x,c))/\tau_{\mathrm{conf}}]_0^1 .
\end{equation}
Disabled checks are omitted. The local risk is
\begin{equation}
r_{\mathrm{local}}(x,c)=1-\min_j s_j .
\end{equation}
Thus a single severe evidence failure is sufficient to raise local risk.

\begin{figure}[H]
\centering
\includegraphics[width=0.82\linewidth]{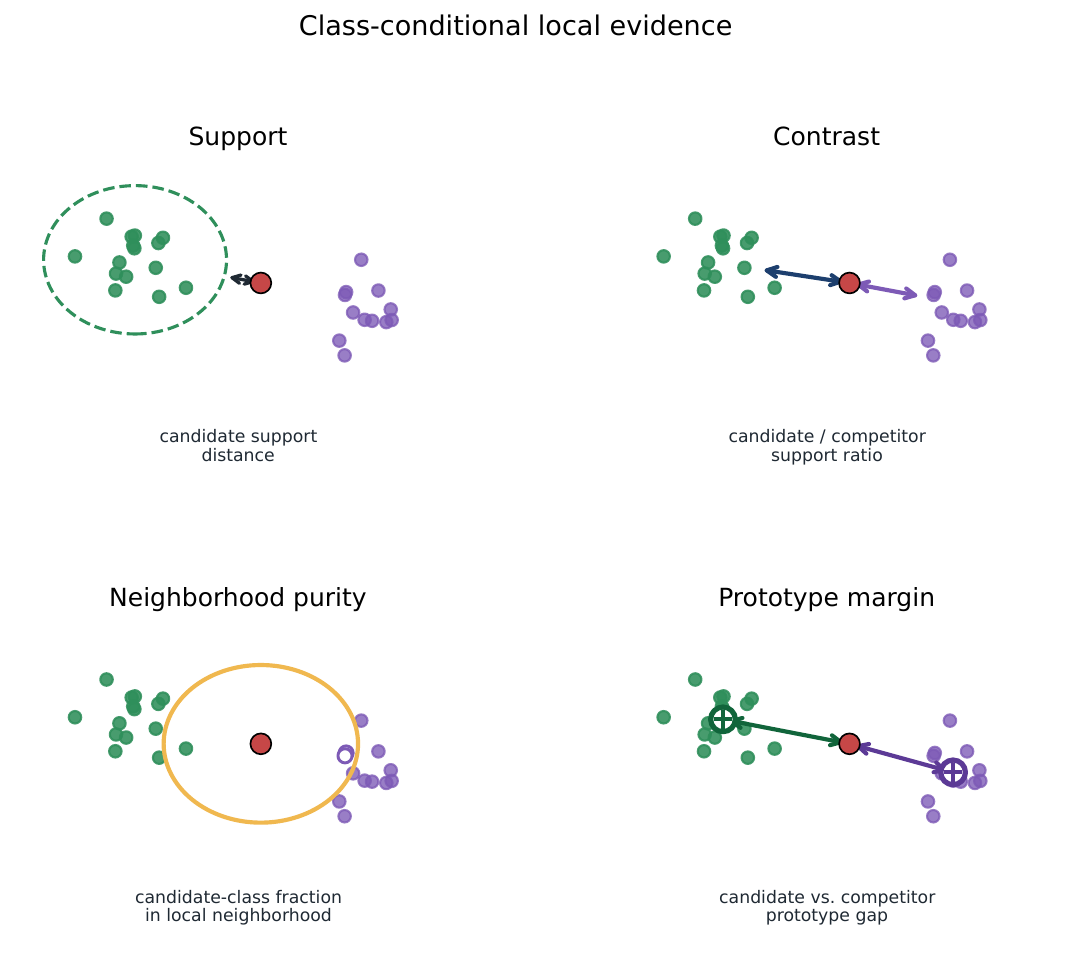}
\caption{Geometric intuition for the four local evidence checks. The main paper describes these as necessary conditions for known-class acceptance; this figure provides the corresponding 2D schematic.}
\label{fig:supp-local-evidence}
\end{figure}

\section{Local Evidence Ablation Detail}

Table \ref{tab:supp-local-ablation} records the Aircraft local-evidence ablations summarized in the main paper.

\begin{table}[H]
\centering
\small
\begin{tabular}{lrr}
\hline
Method & KRR & HC-FKAR \\
\hline
linear MSP & 0.475 & 0.869 \\
EGUR-min (S+C+P+M) & 0.698 & 0.241 \\
w/o support gate & 0.637 & 0.619 \\
support only & 0.171 & 0.537 \\
w/o contrast+purity & 0.586 & 0.320 \\
w/o margin gate & 0.646 & 0.289 \\
global calibration & 0.647 & 0.577 \\
EGUR-A ($\alpha=0.6$) & 0.698 & \textbf{0.071} \\
\hline
\end{tabular}
\caption{Aircraft local-evidence ablation details. EGUR-min denotes support, contrast, purity, and margin before adaptive residual weighting.}
\label{tab:supp-local-ablation}
\end{table}

\section{Sensitivity and Operating-Point Diagnostics}

\begin{figure}[H]
\centering
\includegraphics[width=\linewidth]{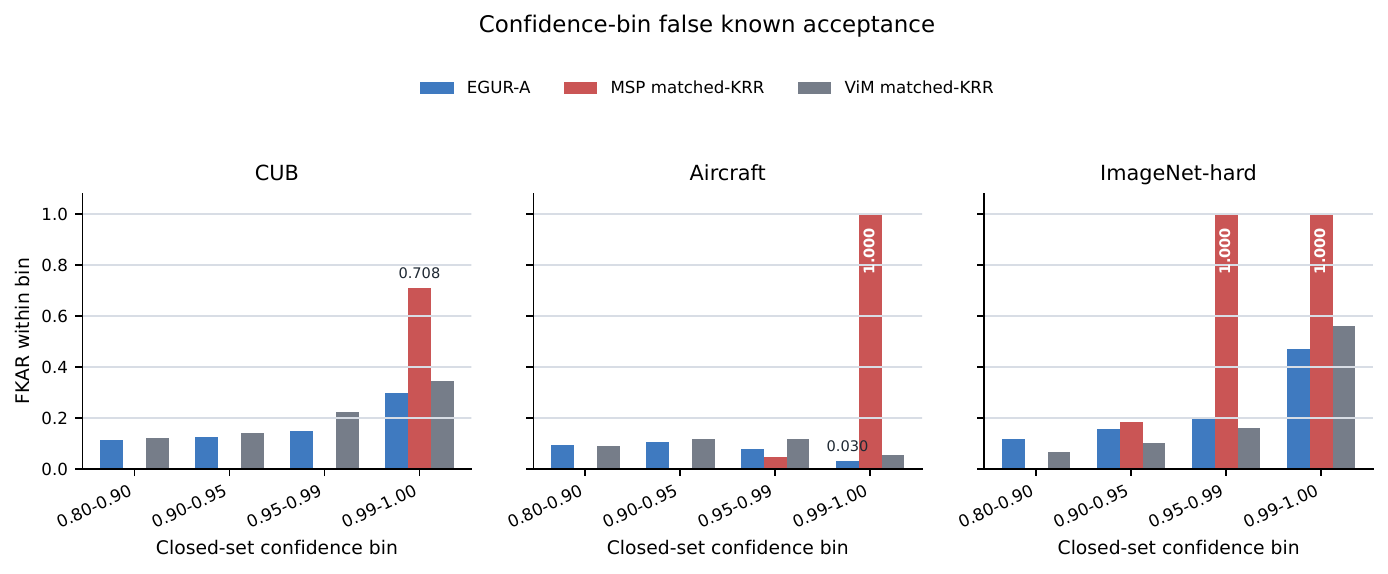}
\caption{Complete confidence-bin FKAR comparison for EGUR-A, MSP, and ViM at the fair operating points. This figure should be interpreted as a high-confidence scalar-threshold failure analysis, not as an all-bin dominance claim for EGUR-A over ViM.}
\label{fig:supp-confidence-bin-full}
\end{figure}

\begin{figure}[H]
\centering
\includegraphics[width=\linewidth]{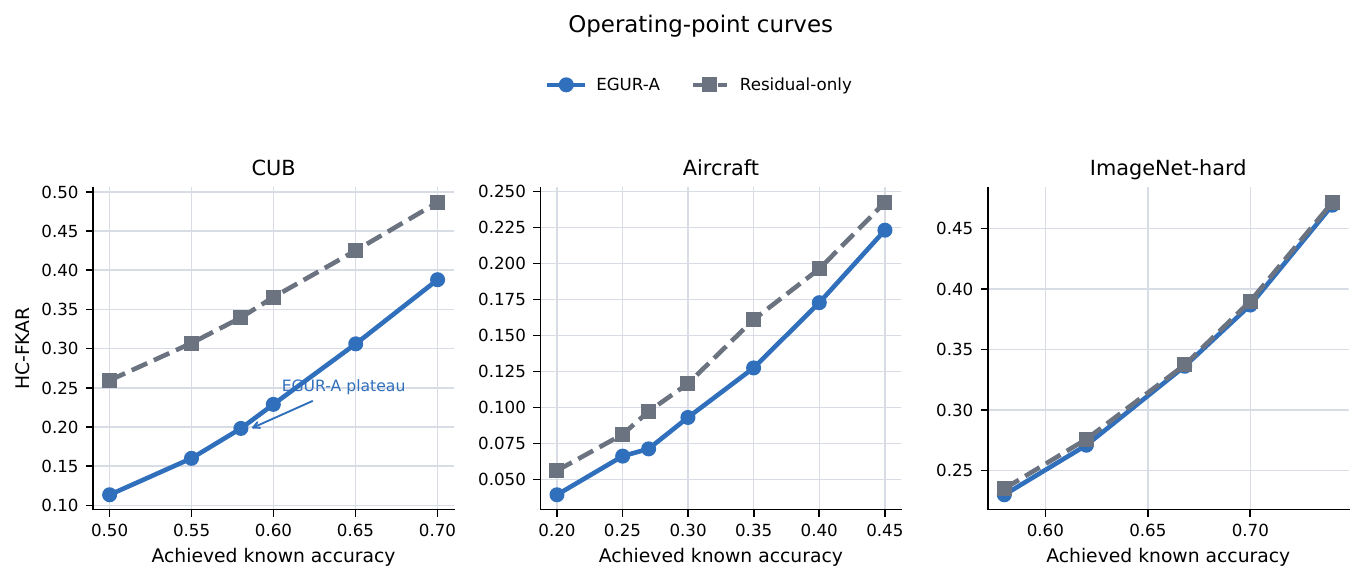}
\caption{Operating-point curves for EGUR-A and residual-only risk. CUB shows a large local-evidence advantage; Aircraft preserves an EGUR-A HC-FKAR advantage across the practical range; ImageNet-hard is residual-saturated, so the curves nearly overlap.}
\label{fig:supp-operating-curves}
\end{figure}

The main paper reports the HC-FKAR@t sensitivity curve. Figure \ref{fig:supp-confidence-bin-full} gives the complete confidence-bin comparison including ViM, and Figure \ref{fig:supp-operating-curves} gives the three-dataset operating-point curves.

\begin{table}[H]
\centering
\small
\begin{tabular}{lrr}
\hline
Method & FKAR mean$\pm$std & HC-FKAR mean$\pm$std \\
\hline
EGUR-A $\alpha=0.2$ & 0.1266$\pm$0.0010 & 0.3362$\pm$0.0032 \\
ViM matched-KRR & 0.1112$\pm$0.0010 & 0.3465$\pm$0.0034 \\
MSP matched-KRR & 0.2405$\pm$0.0015 & 0.8468$\pm$0.0029 \\
\hline
\end{tabular}
\caption{Cached class-stratified bootstrap on ImageNet-hard, 1000 repeats. The 980 hard unknown classes are fixed; each repeat samples 50 images with replacement from the cached 50 images in each class. This estimates within-cache sampling variance and is not a new raw-image resampling experiment.}
\label{tab:supp-imagenet-bootstrap}
\end{table}

\section{Full Default-Workpoint Baselines}

Tables \ref{tab:supp-cub-default}, \ref{tab:supp-aircraft-default}, and \ref{tab:supp-imagenet-default} expand the main paper's default-workpoint stress test. These are not matched-KRR comparisons; they show the natural/default operating point of each scalar method.

\begin{table}[H]
\centering
\small
\begin{tabular}{lrrrrrr}
\hline
Method & Known Acc & KRR & FKAR & HC-FKAR & AUROC & FPR95 \\
\hline
MSP & 0.754 & 0.238 & 0.163 & 0.442 & 0.852 & 0.824 \\
Energy & 0.835 & 0.145 & 0.318 & 0.622 & 0.831 & 0.782 \\
MaxLogit & 0.830 & 0.152 & 0.308 & 0.620 & 0.836 & 0.780 \\
GEN & 0.833 & 0.148 & 0.311 & 0.619 & 0.834 & 0.792 \\
ReAct-Energy & 0.836 & 0.145 & 0.318 & 0.621 & 0.832 & 0.779 \\
NNGuide & 0.859 & 0.117 & 0.382 & 0.695 & 0.806 & 0.836 \\
fDBD & 0.867 & 0.115 & 0.480 & 0.823 & 0.822 & 0.733 \\
SCALE-Energy & 0.856 & 0.119 & 0.375 & 0.672 & 0.814 & 0.792 \\
NCI-OOD & 0.896 & 0.056 & 0.944 & 0.963 & 0.444 & 0.973 \\
CADRef & 0.889 & 0.077 & 0.515 & 0.800 & 0.822 & 0.661 \\
ViM & 0.599 & 0.391 & 0.138 & 0.258 & 0.826 & 0.615 \\
\hline
\end{tabular}
\caption{CUB default-workpoint scalar baselines.}
\label{tab:supp-cub-default}
\end{table}

\begin{table}[H]
\centering
\small
\begin{tabular}{lrrrrrr}
\hline
Method & Known Acc & KRR & FKAR & HC-FKAR & AUROC & FPR95 \\
\hline
MSP & 0.495 & 0.475 & 0.311 & 0.869 & 0.648 & 0.804 \\
Energy & 0.667 & 0.221 & 0.769 & 0.982 & 0.555 & 0.952 \\
MaxLogit & 0.651 & 0.247 & 0.717 & 0.980 & 0.565 & 0.951 \\
GEN & 0.655 & 0.240 & 0.734 & 0.980 & 0.575 & 0.919 \\
ReAct-Energy & 0.672 & 0.215 & 0.775 & 0.979 & 0.555 & 0.949 \\
NNGuide & 0.682 & 0.197 & 0.793 & 0.981 & 0.527 & 0.961 \\
fDBD & 0.691 & 0.195 & 0.821 & 0.991 & 0.511 & 0.951 \\
SCALE-Energy & 0.682 & 0.194 & 0.796 & 0.984 & 0.545 & 0.944 \\
NCI-OOD & 0.722 & 0.058 & 0.913 & 0.951 & 0.551 & 0.897 \\
CADRef & 0.724 & 0.116 & 0.880 & 0.988 & 0.527 & 0.822 \\
ViM & 0.150 & 0.837 & 0.030 & 0.029 & 0.653 & 0.769 \\
\hline
\end{tabular}
\caption{FGVC-Aircraft default-workpoint scalar baselines. ViM has low HC-FKAR only at very high KRR.}
\label{tab:supp-aircraft-default}
\end{table}

\begin{table}[H]
\centering
\small
\begin{tabular}{lrrrrrr}
\hline
Method & Known Acc & KRR & FKAR & HC-FKAR & AUROC & FPR95 \\
\hline
MSP & 0.687 & 0.263 & 0.217 & 0.763 & 0.846 & 0.589 \\
Energy & 0.737 & 0.164 & 0.191 & 0.577 & 0.901 & 0.537 \\
MaxLogit & 0.735 & 0.169 & 0.188 & 0.577 & 0.899 & 0.539 \\
GEN & 0.736 & 0.170 & 0.197 & 0.601 & 0.895 & 0.546 \\
ReAct-Energy & 0.738 & 0.163 & 0.192 & 0.578 & 0.901 & 0.537 \\
NNGuide & 0.753 & 0.139 & 0.207 & 0.604 & 0.903 & 0.524 \\
fDBD & 0.751 & 0.150 & 0.210 & 0.611 & 0.897 & 0.555 \\
SCALE-Energy & 0.742 & 0.153 & 0.208 & 0.610 & 0.895 & 0.571 \\
NCI-OOD & 0.778 & 0.055 & 0.981 & 0.980 & 0.387 & 0.975 \\
CADRef & 0.779 & 0.100 & 0.243 & 0.668 & 0.906 & 0.498 \\
ViM & 0.755 & 0.130 & 0.186 & 0.525 & 0.918 & 0.466 \\
\hline
\end{tabular}
\caption{ImageNet-hard default-workpoint scalar baselines.}
\label{tab:supp-imagenet-default}
\end{table}

\paragraph{Matched-KRR scope.}
The main paper reports the full matched-KRR table for linear-probe post-hoc baselines. These rows share the same closed-set linear-probe confidence, so HC-FKAR is measured on the same high-confidence subset. Classical distance and prototype-style baselines use prototype-derived candidate confidence rather than the linear-probe confidence used by the main HC-FKAR tables, so they are reported separately as auxiliary baselines rather than mixed into the main matched-KRR table.

\begin{table}[H]
\centering
\small
\begin{tabular}{llrrrr}
\hline
Dataset & Method & Known Acc & KRR & FKAR & HC-FKAR \\
\hline
CUB & kNN distance & 0.894 & 0.054 & 0.608 & 0.611 \\
CUB & Prototype distance & 0.881 & 0.068 & 0.579 & 0.583 \\
CUB & Diag. Mahalanobis & 0.649 & 0.332 & 0.247 & 0.248 \\
Aircraft & kNN distance & 0.441 & 0.045 & 0.905 & 0.903 \\
Aircraft & Prototype distance & 0.449 & 0.059 & 0.917 & 0.917 \\
Aircraft & Diag. Mahalanobis & 0.301 & 0.653 & 0.140 & 0.133 \\
ImageNet-hard & kNN distance & 0.759 & 0.050 & 0.425 & 0.464 \\
ImageNet-hard & Prototype distance & 0.762 & 0.053 & 0.289 & 0.331 \\
\hline
\end{tabular}
\caption{Additional classical distance and prototype-style baselines at default workpoints. These methods use prototype-derived candidate confidence rather than the linear-probe confidence used by the main HC-FKAR tables, so they are reported separately.}
\label{tab:supp-distance-baselines}
\end{table}

\section{Aggregate-Metric Sanity Check}

HC-FKAR is the primary metric because it isolates high-confidence near-known unknowns. Aggregate metrics are still useful as sanity checks. Table \ref{tab:supp-aggregate-cub} records the CUB numbers used in the main paper's metric discussion.

\begin{table}[H]
\centering
\small
\begin{tabular}{lrrr}
\hline
Method & AUROC & FPR95 & HC-FKAR \\
\hline
MSP & 0.852 & 0.824 & 0.442 \\
ViM & 0.826 & 0.615 & 0.258 \\
EGUR-A & 0.806 & 0.491 & 0.198 \\
\hline
\end{tabular}
\caption{CUB aggregate metrics and high-confidence metric. EGUR-A improves FPR95 and HC-FKAR, while AUROC reflects that EGUR-A is not optimized for the full unknown distribution ranking.}
\label{tab:supp-aggregate-cub}
\end{table}

\section{Evidence-Weight Selection Evidence}

EGUR-A selects the evidence weight $\alpha$ without unknown validation data. The rule compares known-sample stability of local and residual risk. If residual risk has lower coefficient of variation than local risk, EGUR-A selects the residual-dominant endpoint $\alpha=0.2$; otherwise it selects the known-accuracy-maximizing evidence weight after known-only threshold calibration and moves one grid step toward residual evidence. Table \ref{tab:supp-auto-alpha} records the selected evidence weight and resulting main operating-point metrics.

\begin{table}[H]
\centering
\small
\begin{tabular}{lrrr}
\hline
Dataset & $\mathrm{CV}_{\mathrm{local}}$ & $\mathrm{CV}_{\mathrm{res}}$ & $\alpha$ \\
\hline
CUB & 0.182 & 0.231 & 0.8 \\
Aircraft & 0.146 & 0.429 & 0.6 \\
ImageNet-hard & 0.371 & 0.227 & 0.2 \\
Stanford Cars & 0.250 & 0.309 & 0.8 \\
\hline
\end{tabular}
\caption{Known-sample stability statistics used by the unknown-free evidence-weight rule. Stanford Cars is not used to design the rule; it is a held-out check.}
\label{tab:supp-alpha-cv}
\end{table}

\begin{table}[H]
\centering
\small
\begin{tabular}{lrrrrrr}
\hline
Dataset & $\alpha$ & $\alpha_{\mathrm{KA}}$ & Known Acc & KRR & FKAR & HC-FKAR \\
\hline
CUB & 0.8 & 1.0 & 0.580 & 0.411 & 0.116 & 0.198 \\
Aircraft & 0.6 & 0.8 & 0.270 & 0.698 & 0.088 & 0.071 \\
ImageNet-hard & 0.2 & 0.8 & 0.668 & 0.238 & 0.127 & 0.336 \\
Stanford Cars & 0.8 & 1.0 & 0.667 & 0.320 & 0.180 & 0.380 \\
\hline
\end{tabular}
\caption{Automatic unknown-free evidence-weight selection. $\alpha_{\mathrm{KA}}$ is the evidence weight that maximizes known accuracy after calibrating the threshold to the target KRR on known samples; EGUR-A applies the fixed residual-direction correction described in the main paper.}
\label{tab:supp-auto-alpha}
\end{table}

\begin{table}[H]
\centering
\small
\begin{tabular}{lrrrrr}
\hline
Dataset & Auto $\alpha$ & Best $\alpha$ & Auto HC & Best HC & Gap \\
\hline
CUB & 0.8 & 1.0 & 0.198 & 0.196 & 0.002 \\
Aircraft & 0.6 & 0.6 & 0.071 & 0.071 & 0.000 \\
ImageNet-hard & 0.2 & 0.2 & 0.336 & 0.336 & 0.000 \\
Stanford Cars & 0.8 & 0.8 & 0.380 & 0.380 & 0.000 \\
\hline
\end{tabular}
\caption{Diagnostic comparison with the best fixed evidence weight by HC-FKAR. This table uses unknown labels only for diagnosis; automatic evidence-weight selection does not use them.}
\label{tab:supp-auto-alpha-gap}
\end{table}

\begin{table}[H]
\centering
\small
\begin{tabular}{lrrrrr}
\hline
Dataset & $\alpha$ & Known Acc & KRR & FKAR & HC-FKAR \\
\hline
CUB & 0.2 & 0.570 & 0.411 & 0.241 & 0.287 \\
CUB & 0.4 & 0.572 & 0.411 & 0.204 & 0.268 \\
CUB & 0.6 & 0.575 & 0.411 & 0.160 & 0.228 \\
CUB & 0.8 & 0.580 & 0.411 & 0.116 & 0.198 \\
CUB & 1.0 & 0.584 & 0.411 & 0.088 & 0.196 \\
Aircraft & 0.2 & 0.258 & 0.698 & 0.121 & 0.081 \\
Aircraft & 0.4 & 0.262 & 0.698 & 0.105 & 0.076 \\
Aircraft & 0.6 & 0.270 & 0.698 & 0.088 & 0.071 \\
Aircraft & 0.8 & 0.279 & 0.698 & 0.090 & 0.110 \\
Aircraft & 1.0 & 0.264 & 0.716 & 0.142 & 0.228 \\
ImageNet-hard & 0.2 & 0.668 & 0.238 & 0.127 & 0.336 \\
ImageNet-hard & 0.4 & 0.678 & 0.238 & 0.131 & 0.368 \\
ImageNet-hard & 0.6 & 0.687 & 0.238 & 0.139 & 0.406 \\
ImageNet-hard & 0.8 & 0.695 & 0.238 & 0.152 & 0.449 \\
ImageNet-hard & 1.0 & 0.690 & 0.254 & 0.180 & 0.510 \\
\hline
\end{tabular}
\caption{Fixed evidence-weight sweep used only as a diagnostic. Known-only selection tends to favor higher local-evidence weight on Aircraft and ImageNet-hard, which improves known coverage but worsens HC-FKAR because near-known unknowns that already enter a known class's attraction basin can also receive favorable local evidence. This motivates the fixed residual-direction correction.}
\label{tab:supp-alpha-grid}
\end{table}

\section{Naive Fusion Additional Points}

The main paper reports matched-KRR naive fusion. Table \ref{tab:supp-naive-pareto} reports the strongest naive-fusion point under a KRR cap of EGUR-A KRR + 0.05. These rows are supplementary and should not be compared directly with the main matched-KRR rows because KRR differs.

\begin{table}[H]
\centering
\small
\begin{tabular}{lrrrrr}
\hline
Dataset & $\beta$ & Known Acc & KRR & FKAR & HC-FKAR \\
\hline
CUB & 0.4 & 0.534 & 0.461 & 0.061 & 0.165 \\
Aircraft & 0.2 & 0.245 & 0.748 & 0.049 & 0.113 \\
ImageNet-hard & 0.0 & 0.620 & 0.288 & 0.108 & 0.276 \\
\hline
\end{tabular}
\caption{Naive local-residual fusion Pareto points under the KRR cap.}
\label{tab:supp-naive-pareto}
\end{table}

\section{Negative Results}

\paragraph{EGUR-V.}
Residual-priority vetoes verified that residual evidence can be powerful, but the rule was not a viable unified main method. On Aircraft, one selected point reached HC-FKAR 0.005 but required KRR 0.943 and Known Acc 0.052. On CUB, a selected point reached HC-FKAR 0.130 with KRR 0.488 and Known Acc 0.507. ImageNet-hard was acceptable, but not enough to offset the CUB/Aircraft degradation.

\paragraph{Diagonal per-class Mahalanobis.}
Diagonal per-class Mahalanobis did not close the residual-signal gap. On Aircraft near the EGUR main KRR, it reached KRR 0.697 and HC-FKAR 0.372, worse than the local EGUR-min reference point (HC-FKAR 0.241). Pushing HC-FKAR down to 0.149 required KRR 0.789, still weaker than ViM matched-KRR at HC-FKAR 0.097 and KRR 0.698.

\section{Far-OOD Safety}

EGUR-A is designed for near-known unknowns, but it should not degrade far-OOD rejection. In the DTD check, CUB EGUR-A yields DTD-FKAR 0.000, and Aircraft EGUR-A with $\alpha=0.6$ also yields DTD-FKAR 0.000.

\section{Protocol Note}

The ImageNet-hard compact benchmark uses 500 known classes and 49,000 hard unknown samples, with 50 samples per hard unknown class sampled from the official SSB hard split. This is a compact hard split evaluation, not a full ImageNet21K-scale evaluation.